\documentclass[conference]{IEEEtran}
\usepackage{float}

\usepackage[T1]{fontenc}
\usepackage{graphicx}
\usepackage{amsmath,amssymb}
\usepackage{booktabs}
\usepackage{url}
\usepackage[hidelinks]{hyperref} 
\usepackage{tikz}
\usetikzlibrary{arrows.meta,positioning,shapes.geometric,calc}
\usepackage{adjustbox}
\usepackage{float}

\begin{document}
\setlength{\itemsep}{0pt}
\setlength{\parskip}{0pt}

\title{D-SECURE: Dual-Source Evidence Combination for Unified Reasoning in Misinformation Detection}

\author{
\IEEEauthorblockN{Samudi Amarasinghe, Gagandeep Singh, Priyanka Singh}
\IEEEauthorblockA{
School of Information Technology and Electrical Engineering\\
The University of Queensland, Brisbane, Australia\\
samudi.amarasinghe@student.uq.edu.au\\
gagandeep.singh2@student.uq.edu.au\\
priyanka.singh@uq.edu.au
}
}

\maketitle

\begin{abstract}
Multimodal misinformation increasingly mixes realistic image edits with fluent but misleading text, producing persuasive posts that are difficult to verify. Existing systems usually rely on a single evidence source. Content-based detectors identify local inconsistencies within an image and its caption but cannot determine global factual truth. Retrieval-based fact-checkers reason over external evidence but treat inputs as coarse claims and often miss subtle visual or textual manipulations. This separation creates failure cases where internally consistent fabrications bypass manipulation detectors and fact-checkers verify claims that contain pixel-level or token-level corruption.

We present \textbf{D-SECURE}, a framework that combines internal manipulation detection with external evidence-based reasoning for news-style posts. D-SECURE employs rule-based methods and LLM fusion to integrate the HAMMER manipulation detector with the DEFAME retrieval pipeline. DEFAME performs broad verification, and HAMMER analyses residual or uncertain cases that may contain fine-grained edits. Experiments on several dataset samples highlight the complementary strengths of both systems and motivate their fusion. Additionally, we provide a chat feature for D-SECURE's LLM fusion, which supports explainability and in-depth analysis of outputs. 

\end{abstract}

\begin{figure*}[H]
  \includegraphics[width=\textwidth]{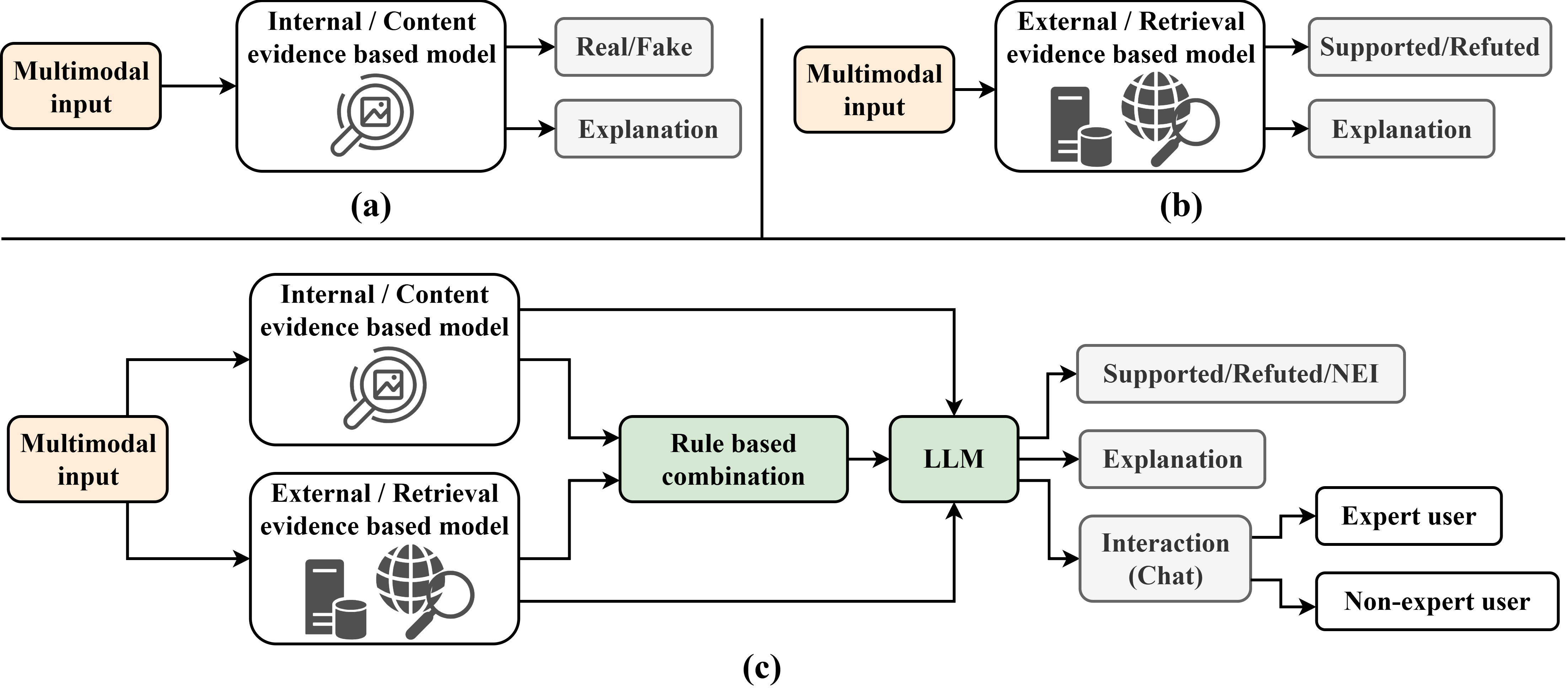}
  \caption{Comparison of D-SECURE with existing work. (a)-(b) Existing systems. (c) D-SECURE}
  \label{fig:diff}
\end{figure*}

\section{Introduction}
\label{sec:intro}

Online platforms have become primary channels through which people consume news, form political opinions, and make health and financial decisions. In this environment, misinformation is not merely an annoyance. It has measurable psychological, social, and economic harms, from vaccine hesitancy and racial discrimination during the COVID-19 pandemic to coordinated political influence operations and deceptive commercial advertising \cite{adams2023misinformation, scheufele2019science, oneil2020australian, nan2022publichealth, meel2020fakenews}. The rise of large language models (LLMs) and generative media has further accelerated the creation of highly realistic but false content \cite{park2025genai, zhou2023synthetic}. Users are routinely exposed to posts that look and read like legitimate news, yet are either locally manipulated or globally incorrect.

A particularly challenging category is \emph{multimodal} misinformation, where images and text are jointly used to convey a message. Empirical work shows that image--text posts achieve higher engagement and are perceived as more credible than text-only posts \cite{li2020picture,akhtar2023multimodalfactcheck, hameleers2020picture}. Social media algorithms, which amplify moral and emotionally charged content, further boost the reach of such posts \cite{mcloughlin2024human,atesgoz2025disinformation}. As a result, multimodal misinformation can spread faster and embed more deeply than traditional single-modality fake news.

\subsection{Internal versus external evidence}

Existing technical work on misinformation detection can be grouped into two broad families.

\paragraph{Content-based manipulation detection.}
One line of research focuses on detecting manipulations \emph{inside} an image--text pair. This includes out-of-context (OOC) pairings where a real image is paired with a misleading caption \cite{luo2021newsclippings,aneja2021cosmos}, as well as local pixel- or token-level edits such as face swaps, sentiment changes, or attribute modifications \cite{shao2023detecting,shao2024detecting,zhu2025multimodal}. HAMMER and the DGM$^4$ benchmark are representative of this direction. They model multimodal reasoning with both global and local views and provide explicit grounding of manipulated regions and tokens \cite{shao2023detecting,shao2024detecting}. These systems are very strong at identifying subtle tampering but are fundamentally limited to the content they see. They do not decide whether an unmanipulated post is factually correct in the real world.

\paragraph{Retrieval-based fact-checking.}
A second line of work abstracts away from local manipulations and focuses on determining the \emph{veracity} of a claim given external evidence. Many systems follow the three-stage pipeline of claim extraction, evidence retrieval, and verdict prediction \cite{akhtar2023multimodalfactcheck}. Early work relied on structured knowledge bases and heterogeneous graphs that connect posts, entities, and external sources \cite{hu2021compare,cao2024multisource}. More recent approaches use retrieval-augmented generation (RAG), where LLMs or MLLMs are guided by retrieved documents, images, or geolocation cues to generate fact-checking verdicts \cite{khaliq2024ragar,braun2024defame}. DEFAME, for example, uses an MLLM with diverse search tools, including web, image, reverse image, and geolocation search, and produces a human-readable report that improves both performance and explainability \cite{braun2024defame}.

These retrieval-based systems can reason about fully fabricated or pristine posts, as long as there is sufficient evidence on the web. However, they often treat the multimodal input as a coarse textual claim and rely on the generative model to implicitly notice manipulations. There is usually no dedicated module for local manipulation detection, and small but crucial edits (for example, swapping a political figure's face) may go unnoticed.

\subsection{Motivation}

We identify an important structural gap in the literature: current systems tend to rely on either internal evidence (content-based manipulation detection) or external evidence (retrieval-based fact-checking), but almost never both in a single, coherent pipeline. This leads to systematic blind spots. Content-based detectors can tell that an image has been manipulated, but cannot say whether an unmanipulated image supports a false narrative. Retrieval-based pipelines can decide whether a claim is supported, refuted, or undecidable, yet they may confidently process content that is already locally corrupted \cite{akhtar2023multimodalfactcheck,shao2023detecting,braun2024defame}.  

\subsection{Contributions}

We address these challenges with D-SECURE, a dual-source framework for multimodal misinformation detection. Concretely, our contributions are:

\begin{enumerate}
  \item We formalise the task of dual-source multimodal misinformation detection that jointly reasons over content-based cues and retrieval-based evidence for news-style image--text posts.
  \item We propose the D-SECURE pipeline, which integrates the HAMMER manipulation detector \cite{shao2023detecting,shao2024detecting} with the DEFAME retrieval-based fact-checker \cite{braun2024defame}. DEFAME acts as a broad-coverage pre-screening stage, while HAMMER provides fine-grained analysis on residual or uncertain cases.
  \item We perform cross-system experiments where DGM$^4$ samples are evaluated with DEFAME and ClaimReview-style fact-checking samples are passed through HAMMER, analysing the strengths and weaknesses of each component and their impact on pipeline design. We further evaluate on other benchmarks to test the generalisability of the pipeline.
  \item We extend explanations across the full pipeline via an LLM-backed chat feature, thereby improving interpretability for downstream users.
\end{enumerate}

\section{Related Work}
\label{sec:related}

We group related work into four categories: multimodal manipulation detection, LLMs and misinformation, retrieval-based fact-checking, and hybrid approaches that combine internal and external evidence.

\subsection{Multimodal manipulation detection}

Early work on multimodal misinformation focused on out-of-context (OOC) manipulations where authentic images are paired with misleading captions. NewsClippings \cite{luo2021newsclippings} automatically constructs OOC examples from news archives, while COSMOS \cite{aneja2021cosmos} uses self-supervised learning over image--caption co-occurrences to detect when a given pairing is inconsistent with typical usage. These methods model global semantic mismatch at the pair level, yet they do not attempt to localise specific pixels or tokens that are responsible for the inconsistency.

The DGM$^4$ benchmark and HAMMER model \cite{shao2023detecting,shao2024detecting} explicitly address local multimodal manipulations. DGM$^4$ defines classes for face swap (FS), face attribute (FA), text swap (TS), and text attribute (TA), as well as their combinations, and provides bounding-box annotations and token labels for manipulated regions. HAMMER uses manipulation-aware contrastive learning with both global and local views, aligning consistent image--text pairs and repelling mismatched or manipulated ones, while also predicting manipulation class, manipulated regions, and manipulated tokens. This line of work gives strong local supervision and explicit grounding, but by design, it operates purely on internal content. It cannot determine whether a pristine sample depicts an event that has ever happened in the world.

Recent datasets such as MFND \cite{zhu2025multimodal} extend multimodal fake news detection with more diverse manipulations and hard negatives. However, they still focus on content-level cues and do not explicitly integrate external factual knowledge beyond simple text-based features.

\subsection{LLMs and Misinformation}
The rapid rise and advancement of generative models such as GPT \cite{openai2023gpt4}, Gemini \cite{team2023gemini} and LLaMA \cite{touvron2023llama} have come alongside several pathways for their misuse. LLMs have conquered the domain and are producing content that is increasingly human-like \cite{bohan2024disinformation}. This content often exploits cognitive biases making individuals more susceptible to being misinformed \cite{park2025genai}. These developments have segued into the trend of LLM-generated misinformation content, such as deepfakes or fake news \cite{park2025genai, pelrine2023towards}. In addition to this, LLMs hold the risk of unintentionally generating misinformation and disinformation as a result of hallucinations \cite{chen2024combat, huang2025survey, massenon2025myphone, sun2024ai}. The misinformation generated by these models is often perceived to be more challenging to detect, incorporating more deceptive styles than human-created content. Despite this, with the spirit of fighting fire with fire, using LLMs to fight LLMs has been an avenue for research \cite{chen2024combat, pelrine2023towards, bohan2024disinformation}. This is desirable as these models are considered human-friendly and are capable of providing explanations to facilitate better understanding \cite{chen2024combat}. For this reason, studies such as \cite{tariq2025p2e} incorporate LLMs in chat features to improve explainability of their systems. Furthermore, dialogue with LLMs has been shown to be effective in persuading individuals away from misinformation \cite{costello2024dialogue}. However, the challenge of misinformation detection proves to be formidable. \cite{bohan2024disinformation} found that existing detection models, such as RoBERTa, can identify misinformation in human-written datasets and simple LLM-generated misinformation, but struggle to detect more complex interleaved content. Nevertheless, more recent models, especially if provided with CoT (Chain-of-Thought) prompts, have a tendency to perform better \cite{chen2024combat, bohan2024disinformation, huang2023reasoning}. \cite{pelrine2023towards} found that the GPT-4 \cite{openai2023gpt4} model performs well in this task even in datasets past its knowledge cutoff. Furthermore, these models can obtain improved performance when web retrieval is utilised to account for their knowledge cutoffs.

\subsection{Retrieval-based fact-checking}

Retrieval-based fact-checking systems aim to decide whether a claim is supported or refuted by external evidence. Many follow a three-stage pipeline of claim extraction, evidence retrieval, and verdict prediction \cite{akhtar2023multimodalfactcheck}. One line of work uses heterogeneous graphs to represent posts, entities, and external knowledge bases. Hu et al.\ \cite{hu2021compare} construct a directed heterogeneous graph that includes news articles and entities from external knowledge bases, then apply graph neural networks to classify fake news. Cao et al.\ \cite{cao2024multisource} propose multi-source knowledge enhanced graph attention networks that combine multimodal evidence with external knowledge, improving veracity prediction by reasoning over cross-modal consistency and external indicators.

More recently, retrieval-augmented generation has become popular. RAGAR \cite{khaliq2024ragar} uses an LLM to verbalise multimodal input, generate clarifying questions, perform iterative retrieval, and produce fact-checking verdicts. DEFAME \cite{braun2024defame} uses an MLLM directly on multimodal inputs and augments it with a diverse set of search tools, including web search, image search, reverse image search, and geolocation. DEFAME also introduces a structured, human-readable report that records retrieved evidence and intermediate reasoning steps, which serves both as in-context guidance for the MLLM and as an explanation for human users. Furthermore, to support retrieval-based fact-checking several benchmarks such as MOCHEG \cite{yao2023end} and ClaimReview2024+ \cite{braun2024defame} have been introduced. 

The aforementioned systems demonstrate the value of rich external evidence. However, they typically have only weak access to internal manipulation cues. RAGAR verbalises images into text, which can lose fine-grained visual details. DEFAME relies on the MLLM's implicit ability to notice mismatches but does not include a dedicated manipulation detector or grounding module. As a result, they may accept or refute claims based on evidence that matches a clean version of the content, while overlooking subtle local tampering.

\subsection{Hybrid internal / external approaches and remaining gaps}

Some graph-based methods attempt to fuse internal and external cues by including both content features and external knowledge in a single representation \cite{hu2021compare,cao2024multisource}. However, they usually work with relatively coarse, document-level features and do not operate at the level of token- or region-level manipulation. Conversely, content-based manipulation detectors, including HAMMER, are powerful at local reasoning but do not access external knowledge sources \cite{shao2023detecting,shao2024detecting}.

To our knowledge, there is no existing system that:

\begin{itemize}
  \item incorporates a state-of-the-art multimodal manipulation detector with explicit grounding;
  \item integrates this detector into a retrieval-based, multimodal fact-checking pipeline that operates over diverse web and image evidence; and
  \item exposes a unified, human-readable explanation that integrates both external evidence and internal manipulation cues.
\end{itemize}

This gap is precisely where D-SECURE is positioned. By combining DEFAME and HAMMER in a principled dual-source architecture, we target both global factuality and local manipulation while maintaining interpretability.

\subsection{Comparison of related methods}

Table~\ref{tab:related-comparison} summarises key properties of representative systems and highlights where our framework differs.

\begin{table*}[t]
\centering
\caption{Comparison of representative approaches to multimodal misinformation detection and fact-checking. ``Internal'' denotes explicit content-based manipulation reasoning. ``External'' denotes use of retrieved web or knowledge-base evidence. ``Local'' indicates explicit localisation of manipulations at region or token level.}
\label{tab:related-comparison}
\resizebox{\linewidth}{!}{
\begin{tabular}{lcccc}
\toprule
Method & Input & Internal & External & Local \\
\midrule
NewsClippings \cite{luo2021newsclippings} 
  & Image--text & OOC consistency & None & No \\
COSMOS \cite{aneja2021cosmos} 
  & Image--text & Self-supervised co-occurrence & None & No \\
DGM$^4$ / HAMMER \cite{shao2023detecting,shao2024detecting} 
  & Image--text & Contrastive reasoning & None & Yes \\
MFND / SDML \cite{zhu2025multimodal} 
  & Image--text & Global content features & None & Limited \\
Hu et al.\ \cite{hu2021compare} 
  & Text & Content features & KB graphs & No \\
Cao et al.\ \cite{cao2024multisource} 
  & Image--text & Multimodal GNN & KB graphs & Limited \\
RAGAR \cite{khaliq2024ragar} 
  & Image--text & Verbalised content & Web search & No explicit module \\ 
DEFAME \cite{braun2024defame} 
  & Image--text & MLLM reasoning & Web, image, geo & No explicit module \\
\textbf{D-SECURE (ours)} 
  & Image--text & HAMMER + DEFAME & Diverse web tools (DEFAME) & Yes \\
\bottomrule
\end{tabular}
}
\end{table*}

As the table shows, D-SECURE is designed to explicitly occupy the currently missing corner: a dual-source multimodal pipeline that uses both a specialised manipulation detector and a retrieval-based fact-checker with rich external evidence, while providing local grounding and global verdicts.

\section{Methodology}
\label{sec:method}

We now describe the D-SECURE framework. Our goal is to integrate HAMMER and DEFAME into a single pipeline that can utilise both internal and external evidence when judging whether a multimodal post constitutes misinformation.

\subsection{Problem formulation}

Let $x = (I, T)$ denote an image--text pair, where $I$ is an image and $T$ is its associated caption or headline. The task is to assign $x$ a misinformation label $y \in \{\text{supported}, \text{refuted}, \text{not-enough-information}\}$, together with optional manipulation labels $m$ that indicate whether $x$ contains local manipulations (for example FS, FA, TS, TA and their combinations) and where they occur in the image and text.

D-SECURE decomposes this problem into two stages:

\begin{enumerate}
  \item a retrieval-based fact-checking stage that focuses on global veracity based on external evidence; and
  \item a content-based manipulation detection stage that focuses on local tampering in the multimodal content itself.
\end{enumerate}

The final prediction is obtained by combining signals from both stages through a rule-based fusion. Figure 2 summarises all types of samples that we cover in this paper. This rule-based fusion is then combined with the DEFAME and HAMMER outputs to prompt an LLM for a verdict and explanation. This provides a fused representation of the DEFAME, HAMMER and D-SECURE rule-based outputs. To enable understanding for all types of users, inspired by \cite{tariq2025p2e}, we enable a chat feature where the LLM can be interrogated on its verdict.

\begin{figure*}[t] \centering \includegraphics[width=\linewidth]{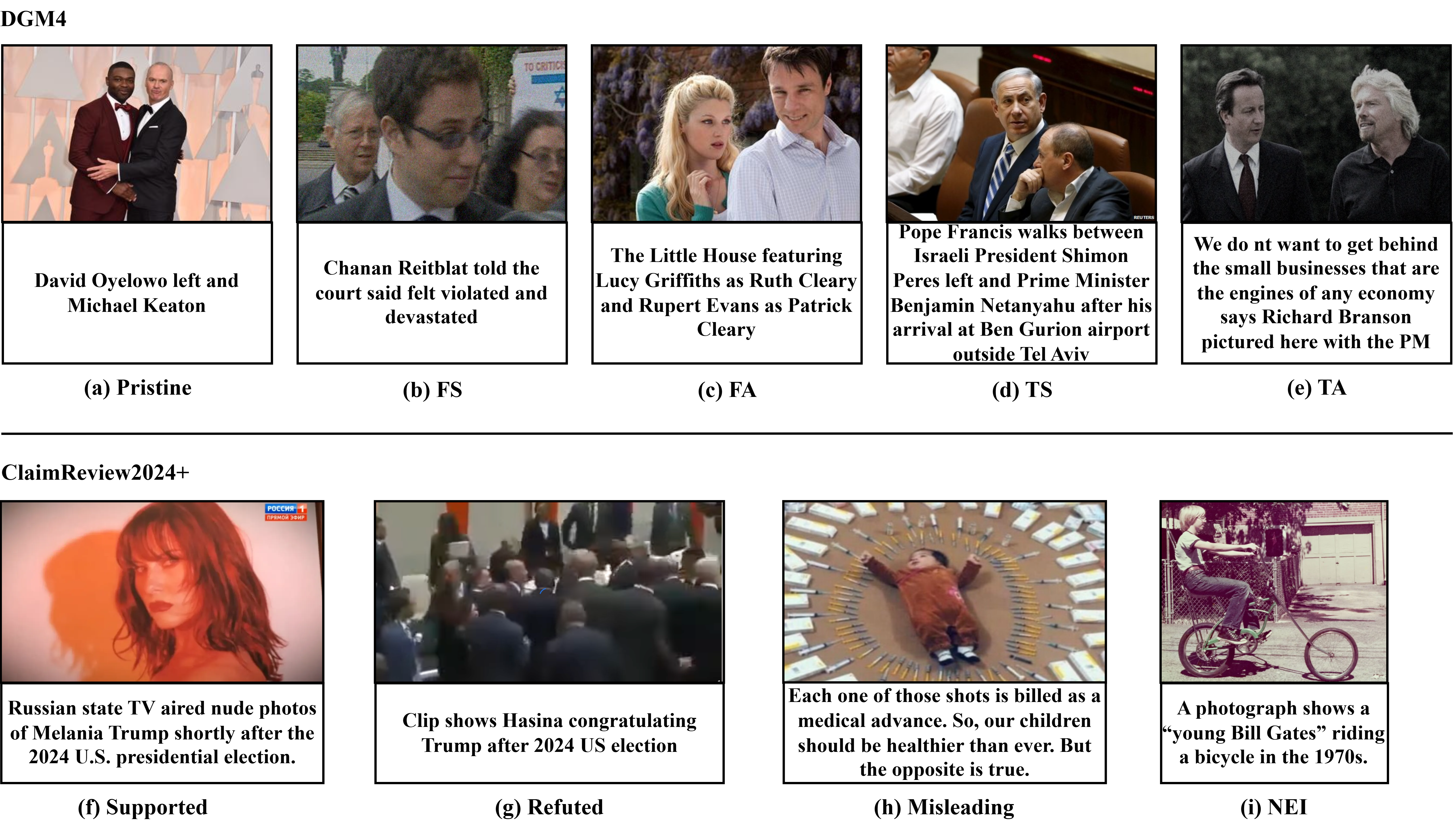} \caption{
\textbf{DGM\textsuperscript{4} and ClaimReview2024+ examples.}
DGM\textsuperscript{4} includes (a) pristine image–text pairs and four fine-grained local manipulation types: (b) Face Swap (FS), (c) Face Attribute (FA), (d) Text Swap (TS), and (e) Text Attribute (TA), as well as their mixed combinations (FS+TS, FS+TA, FA+TS, FA+TA).  
ClaimReview2024+ represents global factuality labels with (f) Supported, (g) Refuted, (h) Misleading and (i) Not-Enough-Information (NEI) cases.
}
\label{fig:datasets} \end{figure*}

\subsection{System components}

\subsubsection{HAMMER}

HAMMER is a hierarchical multimodal manipulation reasoning transformer trained on the DGM$^4$ dataset \cite{shao2023detecting,shao2024detecting}. It takes as input an image and a caption and outputs:

\begin{itemize}
  \item a DGM$^4$ class label (pristine, FS, FA, TS, TA, and their combinations);
  \item a binary real/fake label;
  \item token-level labels for manipulated text positions; and
  \item bounding boxes for manipulated image regions.
\end{itemize}

\subsubsection{DEFAME}

DEFAME is a retrieval-based fact-checking pipeline built around an MLLM backbone and a set of search tools \cite{braun2024defame}. It accepts multimodal inputs and produces:

\begin{itemize}
    \item a verdict in $\{\text{supported}, \text{refuted},$\\
      $\text{not-enough-information (NEI)}\}$ (default verdicts);
  \item a structured, human-readable report that records retrieved evidence, intermediate reasoning, and the final decision.
\end{itemize}

DEFAME can be configured with different web and image search providers. In our experiments, we follow one of the original designs with a free MLLM and paid tools such as Google and dedicated image search, where available \cite{braun2024defame}. BeautifulSoup is used to scrape and normalise retrieved pages before they are passed to the MLLM.

We adopt the configuration where DEFAME acts as a pre-processing step and HAMMER is applied to residual cases. The motivation is twofold. First, DEFAME is designed to capture a broad range of misinformation phenomena by reasoning over external evidence, including fully fabricated or globally false claims for which no local manipulation is required. Second, preliminary experiments where DEFAME is applied to DGM$^4$ samples show that it performs reliably when it refutes a claim (it never refuted truly pristine samples) but struggles with subtle local manipulations, especially in free-tool settings. This suggests that its refuted predictions can be trusted as strong global signals, while additional local analysis is beneficial for uncertain or apparently supported cases.

Formally, given input $x$, D-SECURE first runs DEFAME:

\begin{equation}
  (y^{\text{def}}, r) = f_{\text{DEFAME}}(x),
\end{equation}

where $y^{\text{def}}$ is the fact-checking verdict and $r$ is the associated reasoning report. We then apply the following logic:

\begin{itemize}
  \item If $y^{\text{def}} = \text{refuted}$, we output a final verdict of \textit{refuted}, together with $r$ as the explanation and pass the sample to HAMMER to check whether localised edits caused refutation or purely external evidence and assist explainability.
  \item If $y^{\text{def}} \in \{\text{supported}, \text{not-enough-information}\}$, we pass $x$ to HAMMER for local analysis.
\end{itemize}

HAMMER then produces manipulation labels and localisation:

\begin{equation}
  (m^{\text{ham}}, b^{\text{ham}}) = f_{\text{HAMMER}}(x),
\end{equation}

where $m^{\text{ham}}$ captures manipulation class and token-level labels, and $b^{\text{ham}}$ are bounding boxes for image manipulations. We additionally test the reverse order of the pipeline in our experiments.

\subsection{Decision Fusion}

\begin{figure*}[t]
    \centering
    \includegraphics[width=\linewidth]{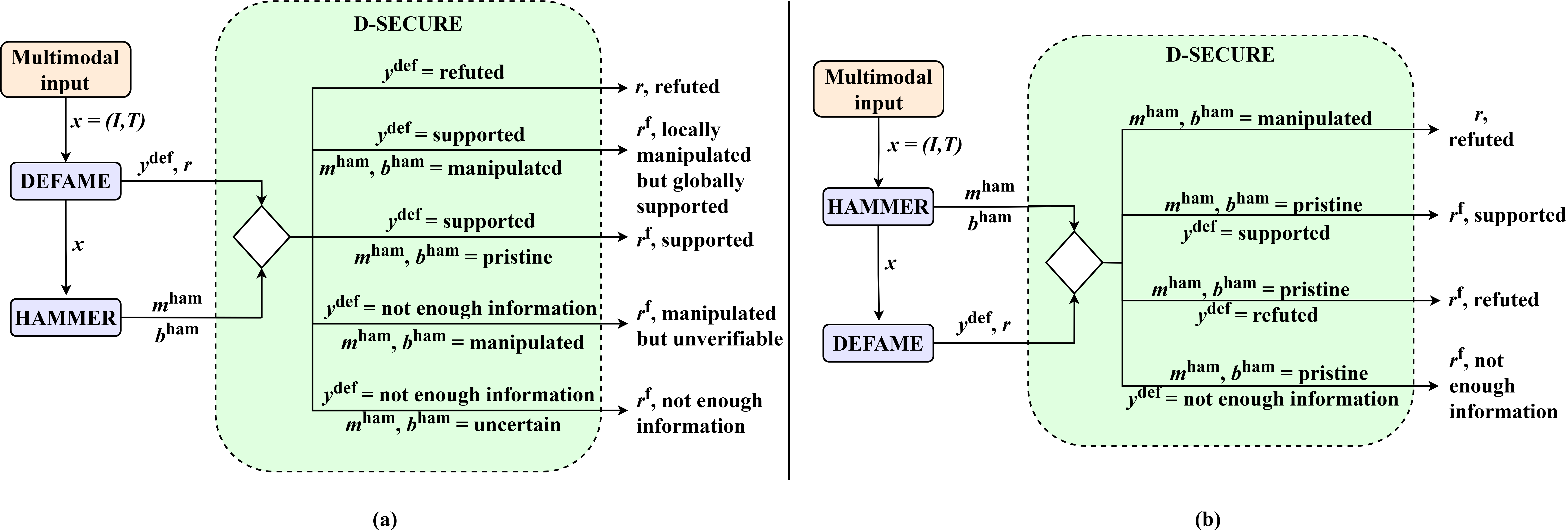}
    \caption{D-SECURE's rule-based scheme for the DEFAME-HAMMER pipeline.}
    \label{fig:d-sec-rules}
\end{figure*}

Figure~\ref{fig:d-sec-rules} summarises D-SECURE’s rule-based fusion strategy.
DEFAME provides the initial global verdict, and HAMMER supplies manipulation
class labels and grounding. Fusion follows a small set of principled cases:
(1) when DEFAME refutes the claim, the final verdict is \textit{refuted};
(2) when DEFAME supports the claim, HAMMER determines whether the post is
\textit{supported} (pristine) or \textit{locally manipulated but globally supported};
and (3) when DEFAME returns \textit{not enough information}, HAMMER distinguishes
between \textit{manipulated but unverifiable} and genuine \textit{NEI}. This rule set
captures the complementarity between global factual evidence and local  
manipulation cues. 

\begin{figure}[t]
    \centering
    \includegraphics[width=\linewidth]{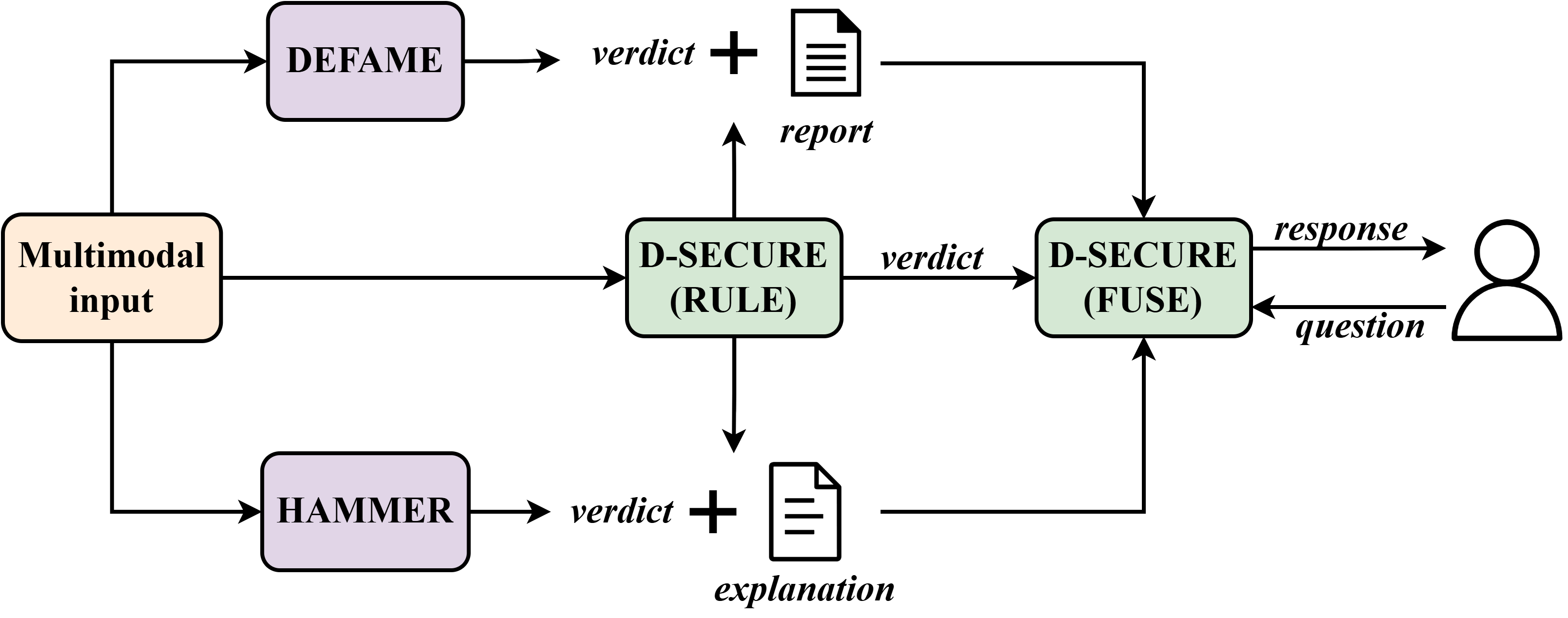}
    \caption{D-SECURE's fusion for the DEFAME-HAMMER pipeline.}
    \label{fig:llm-fusion}
\end{figure}

\begin{figure*}[t]
    \centering
    \includegraphics[width=\linewidth]{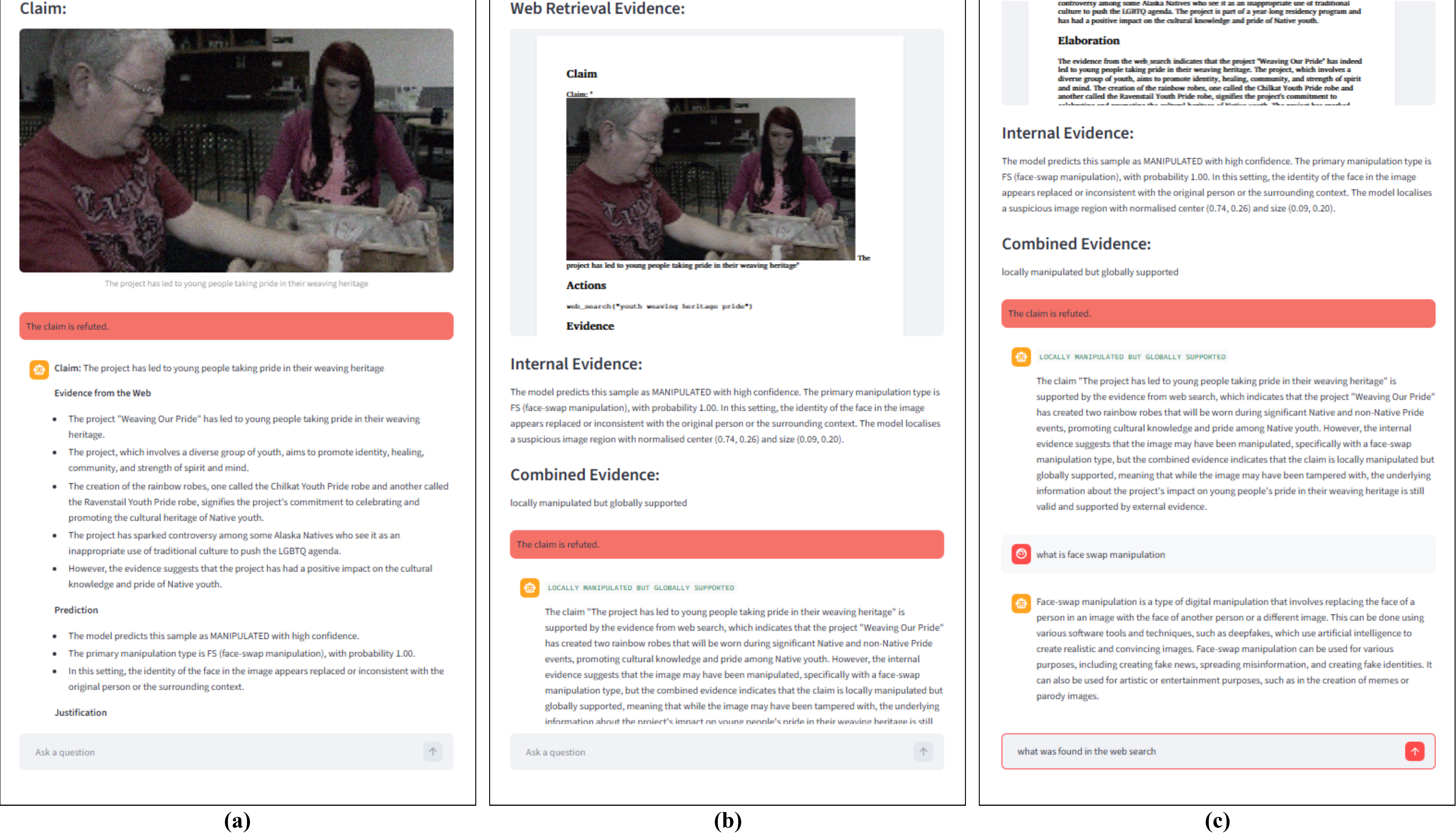}
    \caption{D-SECURE's chat allowing the user to interact and question the LLM. (a) view for non-technical users, (b) view for technical users with evidence, (c) chat interaction between end user and the system.}
    \label{fig:chat-demo}
\end{figure*}

After the verdict for rule-based fusion is obtained, it is provided as input to a LLM alongside DEFAME's report, HAMMER's explanation and the claim (Figure \ref{fig:llm-fusion}). The LLM is prompted to use these three components to make a judgement about the claim's veracity. This judgement involves choosing a verdict out of the same pool as the rule-based method, and providing an explanation for its choice. This provides a fused representation for all the components of the pipeline. 





\subsection{LLM chat}
To ensure that the fused output is interpretable and digestible by users of all levels of expertise we enable a chat feature with our fusion LLM (Figure \ref{fig:chat-demo}). This allows users to ask clarifying questions and query about all pieces of evidence involved in making a prediction about a claim. This feature is intended to improve user trust and support expert audit of system behavior. We provide options for two levels of users: expert answers and non-expert answers. To ensure that non-expert users can understand the system output, we kept the layout simple. We refrained from including all the raw evidence from the model components (DEFAME, HAMMER and D-SECURE rule-based) to ensure that users were not overloaded with technical terminology (Figure \ref{fig:chat-demo} (a)). Instead we prompt the LLM to break up its response into subsections and explain its output in simple language. Conversely, for the expert users we include all pieces of original evidence to allow them to examine the technical information in detail (Figure \ref{fig:chat-demo} (b)). 




\section{Experiments}
\label{sec:experiments}

We evaluate D-SECURE on two complementary axes: (i) local manipulation analysis on the DGM\textsuperscript{4} benchmark, and (ii) global factual verification on a 300-claim 
ClaimReview2024+ benchmark. This joint evaluation reflects the dual nature of multimodal misinformation, where content may be locally edited, globally fabricated, or both. We additionally evaluate D-SECURE on an extension to DGM\textsuperscript{4}, DGM\textsuperscript{4}+ and two external datasets VERITE and Fakeddit to evaluate the pipeline's generalization capability.

\subsection{Datasets}
We evaluate D-SECURE against five benchmarks, DGM\textsuperscript{4}, ClaimReview2024+, DGM\textsuperscript{4}+, VERITE and Fakeddit. The DGM\textsuperscript{4} dataset consists of image-text claims focused on human-centric news. These claims are categorised into an unmanipulated pristine class and four manipulation classes: Face Swap, Face Attribute, Text Swap, and Text Attribute, which are further combined to make combination manipulation classes (Figure \ref{fig:datasets}). DGM\textsuperscript{4} originally consisted of 230k samples; however for our experiments we randomly selected a subset of 720 samples with 80 claims per category. DGM\textsuperscript{4}+ \cite{singh2025dgm4ext} is an extension to the DGM\textsuperscript{4} dataset which involves claims that include fabricated images with foreground-background mismatches, and manipulated text. The original dataset contained 5000, out of which a sample of 750 was taken for our study. ClaimReview2024+ consists of 300 English claims, which are sampled from fact-checking articles across topics such as climate, health and politics. The claims are categorised into Supported (89), Refuted (129), Misleading (61), and Not Enough Information - NEI (21). This yields a class imbalance with the Refuted class as the majority, contributing to 43\% of the dataset. VERITE \cite{papadopoulos2023verite} consists of 1000 image-text claims, including miscaptioned (338) and out-of-context (324) samples. For our experiments, we took a sample of 697 claims from VERITE. The last dataset that was tested was Fakeddit. Fakeddit \cite{nakamura2020rfakeddit} consists of Reddit posts of which 682,996 were multimodal. Of those multimodal claims we took a subset of 700 consisting of 287 true and 413 fake samples.  

\subsection{Experimental Setup}

\paragraph{Systems.}
\textbf{HAMMER} is a state-of-the-art manipulation detector trained on DGM\textsuperscript{4}.  
\textbf{DEFAME} is a retrieval-augmented fact-checking pipeline using a Llava-based MLLM backbone.  
\textbf{D-SECURE (ours)} injects HAMMER’s manipulation-aware summary (binary flag, class, grounded inconsistencies) into DEFAME. D-SECURE produces five labels but these are collapsed to the standard three-way space for evaluation:
Locally Manipulated but Globally Supported (LMGS) $\rightarrow$ Refuted,  
Manipulated but Unverifiable (MBU) $\rightarrow$ Refuted.

\paragraph{Metrics.}
For DGM\textsuperscript{4}, DGM\textsuperscript{4}+, VERITE and Fakeddit we report binary accuracy (ACC). We evaluate the \emph{NEI} predictions as correct, as we consider it important for models to recognise and acknowledge uncertainty, as opposed to making guesses.
Since ClaimReview2024+ uses the same three-way labels as D-SECURE we use 3-way accuracy under three evaluation rules: 
(1) \textit{Strict factuality}, where predictions must match the gold label exactly; 
(2) \textit{Manipulation-aware}, where predictions of \{\emph{Refuted}, \emph{LMGS}, \emph{MBU}\} count as correct for gold Refuted; and 
(3) \textit{Intervention-aware}, where \{\emph{Refuted}, \emph{LMGS}, \emph{MBU}\} are considered correct for gold Refuted or NEI. 
These rules reflect increasingly realistic settings where a practical detector should flag manipulated or unverifiable content even when perfect external evidence is unavailable.

\subsection{Implementation Details}
We use NVIDIA A100 GPUs to conduct the experiments. The MLLM utilised for DEFAME (llava-hf/llava-onevision-qwen2-7b-ov-hf \cite{liu2023llava}) cost approximately 16GB and the LLM used for D-SECURE (meta-llama/Llama-3.2-11B-Vision-Instruct \cite{meta2024llama3}) cost approximately 43GB. The ClaimReview2024+ dataset cost 115MB, the DGM\textsuperscript{4} dataset cost 30GB, DGM\textsuperscript{4}+ cost 98MB, VERITE cost 73MB and Fakeddit cost 216GB. The HAMMER model consisted of all publicly available components and incurred no monetary cost. The DEFAME model required several search tools for which there are both paid and free versions. To maximise performance, we utilized the paid tools and incurred costs for Google Vision API \cite{googlevision} and Serper API \cite{serper} to enable Reverse Image Search and Google Web Search (including Google Image Search), respectively. In addition to these, Geolocation via GeoCLIP \cite{cepeda2023geoclip} was another tool. To scrape web pages for search we utilised Python's BeautifulSoup.

HAMMER obtained a runtime of approximately 0.42 seconds per claim, DEFAME achieved a runtime of approximately 1.8 minutes per claim, and rule-based D-SECURE obtained negligible time per claim. Fusion-based D-SECURE obtains approximately 3 minutes per claim.

\subsection{Results and Analysis}

\begin{table*}[t]
\centering
\caption{Unified results on DGM\textsuperscript{4} and ClaimReview2024+. All values are computed directly from the submitted outputs.}
\vspace{1mm}
\begin{tabular}{l|c|ccc|c|c|c}
\toprule
 & \textbf{DGM\textsuperscript{4}} & \multicolumn{3}{|c|}{\textbf{ClaimReview2024+}} & \textbf{DGM\textsuperscript{4}+} & \textbf{VERITE} & \textbf{Fakeddit} \\
 \cmidrule{3-5} 
\textbf{Model} &
\textbf{  } &
\textbf{ Strict } &
\textbf{Manip.-Aware } &
\textbf{Interv.-Aware} & & \\
\midrule
HAMMER & 90.69 $\pm$ 0.00 & 44.46 $\pm$ 0.30 & -- & -- & 18.13 $\pm$ 0 & 41.75 $\pm$ 0 & 58.14 $\pm$ 0\\
DEFAME & 64.08 $\pm$ 14.63 & 38.60 $\pm$ 4.56 & 41.2 $\pm$ 4 & 40 $\pm$ 4.96 & 78.24 $\pm$ 5.13 & 62.67 $\pm$ 9.22 & \textbf{79.2 $\pm$ 5.39}\\
\textbf{D-SECURE}\textsubscript{DEF-HAM} & \textbf{93.61 $\pm$ 1.3} & 30.8 $\pm$ 3.51 & 48.93 $\pm$ 1.74 & 51.73 $\pm$ 2.96 & 82 $\pm$ 3.87 & 61.64 $\pm$ 6.58 & 74.66 $\pm$ 4.49\\
\textbf{D-SECURE}\textsubscript{HAM-DEF} & \textbf{93.61 $\pm$ 1.3} & \textbf{47.67 $\pm$ 2.29} & 48.93 $\pm$ 1.74 & 51.73 $\pm$ 2.96 & 82 $\pm$ 3.87 & 61.64 $\pm$ 6.58 & 74.66 $\pm$ 4.49\\
\textbf{D-SECURE}\textsubscript{FUSE} & 82.39 $\pm$ 6.38 & 43.53 $\pm$ 3.29 & \textbf{56.67 $\pm$ 4.07} & \textbf{59.2 $\pm$ 3.4} & \textbf{86.03 $\pm$ 3.29} & \textbf{69.41 $\pm$ 3.13} & 78.06 $\pm$ 2.87\\
\bottomrule
\end{tabular}
\label{tab:main_results}
\end{table*}

Table \ref{tab:main_results} displays the results of all pipeline components over the benchmarks.

\paragraph{Local manipulation (DGM\textsuperscript{4}).}
HAMMER achieves 90.69\% binary accuracy on the DGM\textsuperscript{4} predictions, as expected due to the model being trained on this dataset. DEFAME performs worse than HAMMER with accuracy 64.08\%. Rule-based D-SECURE outperforms its ablated components with a slight improvement at 93.61\%. 

\paragraph{Global factuality (ClaimReview2024+).}
Using strict 3-way matching, DEFAME attains 38.60\% accuracy due to its conservative NEI bias. Rule-based D-SECURE in the HAMMER-DEFAME order obtains the highest strict accuracy of 47.67\%, outperforming both DEFAME and D-SECURE. However, the imbalance of classes in the dataset may have impacted strict accuracy performance.
Under manipulation-aware evaluation, fusion-based D-SECURE reaches 56.67\%, and under intervention-aware evaluation, it reaches 59.2\%, outperforming DEFAME (41.2\% and 40\%, respectively). These gains reflect cases where HAMMER correctly identifies local manipulations or inconsistencies that DEFAME alone cannot detect.

\paragraph{Foreground-Background Inconsistencies (DGM\textsuperscript{4}+).} 
HAMMER obtains the weakest performance on this dataset with an accuracy of 18.13\%. This highlights how local analysis falls short on media that have been generated. Fusion-based D-SECURE achieves the highest performance with 86.03\%. 

\paragraph{Out-of-context (VERITE)} 
Fusion-based D-SECURE achieves the best performance with 69.41\%. This depicts that a learned representation of the three signals is beneficial in detecting out-of-context claims. Overall, DEFAME and rule-based D-SECURE maintained similar accuracy while HAMMER resulted in the lowest.

\paragraph{Social media (Fakeddit).} 
DEFAME achieves the highest score with 79.2\%, though fusion based D-SECURE obtains a comparable score of 78.06\%. Social media such as Reddit have a clear benefit from external evidence for misinformation detection.  

\subsection{Discussion}

The results reveal four central observations.

\paragraph{Local manipulation is not a reliable proxy for truth.}
HAMMER excels at detecting pixel- and token-level edits, yet its strict 3-way factual accuracy on ClaimReview2024+ is only 44.46\%. This aligns with the intuition that globally false claims can be paired with completely pristine images, and conversely manipulated content can accompany factually correct narratives. 

\paragraph{Generated media cannot be verified through local reasoning alone.}
HAMMER performs worst on DGM\textsuperscript{4}+ with an accuracy of 18.13\%, indicating that large-scale local manipulations are difficult to detect. The addition of external evidence in DEFAME and D-SECURE significantly improves performance. 

\paragraph{External evidence alone is brittle under subtle edits.}
DEFAME performs reasonably when sufficient web evidence is available, but it frequently defaults to NEI (as seen in its 64.08\% accuracy on DGM\textsuperscript{4}), and misclassifies manipulated posts as Supported when edits are subtle. This explains its limited strict accuracy of 38.60\%.

\paragraph{Combining internal and external signals yields meaningful gains.}
D-SECURE outperforms DEFAME and HAMMER across the majority of benchmarks. The improvement is most pronounced in manipulation and intervention-aware settings of ClaimReview2024+, where HAMMER’s grounded inconsistencies help DEFAME avoid incorrect endorsements. This supports the central claim of the paper: neither local nor global evidence is sufficient alone, and dual-source fusion is necessary for robust multimodal misinformation detection.

\medskip
\noindent
Overall, these results demonstrate that D-SECURE successfully bridges the gap between manipulation sensitivity and factual verification, leading to more reliable and interpretable judgements in diverse misinformation scenarios. 

\section{Limitations}
\label{secc:Limitations}

Our study has three main limitations.
First, D-SECURE relies on HAMMER's manipulation priors and DEFAME's retrieval coverage, both of which may degrade on domains with unfamiliar entities, poor web presence, or highly compressed imagery. Second, the HAMMER model had weak performance on some datasets, which may have affected D-SECURE's performance. Third, the ClaimReview2024+ sample used in our evaluation is relatively small and skewed toward political and health narratives. 
\section{Conclusion}
\label{sec:Conclusion}

We presented D-SECURE, a dual-source framework that unifies local manipulation detection with external evidence retrieval for multimodal misinformation. By combining HAMMER’s grounded manipulation signals with DEFAME’s fact-checking pipeline, D-SECURE closes a persistent gap between content-level tampering and claim-level veracity. Experiments on DGM\textsuperscript{4}, ClaimReview2024+, DGM\textsuperscript{4}+, VERITE and Fakeddit demonstrate that our system maintains state-of-the-art manipulation detection while improving factuality prediction across multiple evaluation rules. This work highlights the importance of bridging internal and external evidence and provides a foundation for future systems that reason jointly over pixels, tokens, and the broader information ecosystem.

\bibliographystyle{IEEEtran}
\bibliography{references}

@article{adams2023misinformation,
  author  = {Adams, Zoe and Osman, Magda and Bechlivanidis, Christos and Meder, Bj{\"o}rn},
  title   = {(Why) Is Misinformation a Problem?},
  journal = {Perspectives on Psychological Science},
  volume  = {18},
  number  = {6},
  pages   = {1436--1463},
  year    = {2023}
}

@inproceedings{zhou2023synthetic,
  author    = {Zhou, Jiaxin and Zhang, Yiqi and Luo, Qi and Parker, Andrea Grimes and De Choudhury, Munmun},
  title     = {Synthetic Lies: Understanding {AI}-Generated Misinformation and Evaluating Algorithmic and Human Solutions},
  booktitle = {Proceedings of the 2023 {CHI} Conference on Human Factors in Computing Systems},
  year      = {2023},
  pages     = {1--20}
}

@book{oneil2020australian,
  author    = {O'Neil, M. and Jensen, M.},
  title     = {Australian Perspectives on Misinformation},
  publisher = {University of Canberra},
  year      = {2020}
}

@article{scheufele2019science,
  author  = {Scheufele, Dietram A. and Krause, Nicole M.},
  title   = {Science Audiences, Misinformation, and Fake News},
  journal = {Proceedings of the National Academy of Sciences},
  volume  = {116},
  number  = {16},
  pages   = {7662--7669},
  year    = {2019}
}

@article{mcloughlin2024human,
  author  = {McLoughlin, K. L. and Brady, William J.},
  title   = {Human--Algorithm Interactions Help Explain the Spread of Misinformation},
  journal = {Current Opinion in Psychology},
  volume  = {56},
  pages   = {101770},
  year    = {2024}
}

@incollection{atesgoz2025disinformation,
  author    = {Ate{\c{s}}g{\"o}z, K.},
  title     = {The Proliferation of Disinformation in the Information and Communication Age: ``News Literacy'' as a Framework for Critical News Reception},
  booktitle = {Digital Literacy as a Catalyst for Critical Thinking: From Media to Artificial Intelligence},
  editor    = {{\"O}zel, M.},
  publisher = {Springer Nature Switzerland},
  year      = {2025},
  pages     = {39--56}
}

@article{li2020picture,
  author  = {Li, Yan and Xie, Yu},
  title   = {Is a Picture Worth a Thousand Words? An Empirical Study of Image Content and Social Media Engagement},
  journal = {Journal of Marketing Research},
  volume  = {57},
  number  = {1},
  pages   = {1--19},
  year    = {2020}
}

@article{akhtar2023multimodalfactcheck,
  author  = {Akhtar, Mubashara and Schlichtkrull, Michael and Guo, Zhijiang and Cocarascu, Oana and Simperl, Elena and Vlachos, Andreas},
  title   = {Multimodal Automated Fact-Checking: A Survey},
  journal = {arXiv preprint arXiv:2305.13507},
  year    = {2023}
}

@article{luo2021newsclippings,
  author  = {Luo, Geng and Darrell, Trevor and Rohrbach, Anna},
  title   = {Newsclippings: Automatic Generation of Out-of-Context Multimodal Media},
  journal = {arXiv preprint arXiv:2104.05893},
  year    = {2021}
}

@inproceedings{aneja2021cosmos,
  author    = {Aneja, Shivangi and Bregler, Christoph and Nie{\ss}ner, Matthias},
  title     = {COSMOS: Catching Out-of-Context Misinformation with Self-Supervised Learning},
  booktitle = {Proceedings of the IEEE/CVF Conference on Computer Vision and Pattern Recognition},
  year      = {2021}
}

@inproceedings{shao2023detecting,
  author    = {Shao, Rui and Wu, Tianyun and Liu, Ziwei},
  title     = {Detecting and Grounding Multi-Modal Media Manipulation},
  booktitle = {Proceedings of the IEEE/CVF Conference on Computer Vision and Pattern Recognition},
  year      = {2023},
  pages     = {6904--6913}
}

@article{shao2024detecting,
  author  = {Shao, Rui and Wu, Tianyun and Wu, Jiaming and Nie, Liqiang and Liu, Ziwei},
  title   = {Detecting and Grounding Multi-Modal Media Manipulation and Beyond},
  journal = {IEEE Transactions on Pattern Analysis and Machine Intelligence},
  volume  = {46},
  number  = {8},
  pages   = {5556--5574},
  year    = {2024}
}

@article{zhu2025multimodal,
  author  = {Zhu, Y. and Wang, Y. and Yu, Z.},
  title   = {Multimodal Fake News Detection: {MFND} Dataset and Shallow--Deep Multitask Learning},
  journal = {arXiv preprint arXiv:2505.06796},
  year    = {2025}
}

@inproceedings{hu2021compare,
  author    = {Hu, Linmei and Deng, Haoran and Hou, Yuxiao and others},
  title     = {Compare to the Knowledge: Graph Neural Fake News Detection with External Knowledge},
  booktitle = {Proceedings of the 59th Annual Meeting of the Association for Computational Linguistics and the 11th International Joint Conference on Natural Language Processing (Volume 1: Long Papers)},
  year      = {2021},
  pages     = {754--763}
}

@inproceedings{cao2024multisource,
  author    = {Cao, H. and Wei, L. and Zhou, W. and Hu, S.},
  title     = {Multi-source Knowledge Enhanced Graph Attention Networks for Multimodal Fact Verification},
  booktitle = {2024 IEEE International Conference on Multimedia and Expo (ICME)},
  year      = {2024},
  pages     = {1--6}
}

@inproceedings{khaliq2024ragar,
  author    = {Khaliq, Muhammad Ali and Chang, Pao-Yen Chen and Ma, Mengting and Pflugfelder, Benjamin and Mileti{\'{c}}, Filip},
  title     = {RAGAR, Your Falsehood Radar: {RAG}-Augmented Reasoning for Political Fact-Checking using Multimodal Large Language Models},
  booktitle = {Proceedings of the Seventh Fact Extraction and VERification Workshop (FEVER)},
  year      = {2024},
  pages     = {280--296}
}

@article{braun2024defame,
  author  = {Braun, Tobias and Rothermel, Michael and Rohrbach, Marcus and Rohrbach, Anna},
  title   = {DEFAME: Dynamic Evidence-based Fact-checking with Multimodal Experts},
  journal = {arXiv preprint arXiv:2412.10510},
  year    = {2024}
}

@article{chen2024combat,
author = {Chen, Canyu and Shu, Kai},
title = {Combating misinformation in the age of LLMs: Opportunities and challenges},
journal = {AI Magazine},
volume = {45},
number = {3},
pages = {354-368},
doi = {https://doi.org/10.1002/aaai.12188},
url = {https://onlinelibrary.wiley.com/doi/abs/10.1002/aaai.12188},
eprint = {https://onlinelibrary.wiley.com/doi/pdf/10.1002/aaai.12188},
year = {2024}
}

@inproceedings{yao2023end,
author = {Yao, Barry Menglong and Shah, Aditya and Sun, Lichao and Cho, Jin-Hee and Huang, Lifu},
title = {End-to-End Multimodal Fact-Checking and Explanation Generation: A Challenging Dataset and Models},
year = {2023},
isbn = {9781450394086},
publisher = {Association for Computing Machinery},
address = {New York, NY, USA},
url = {https://doi.org/10.1145/3539618.3591879},
doi = {10.1145/3539618.3591879},
booktitle = {Proceedings of the 46th International ACM SIGIR Conference on Research and Development in Information Retrieval},
pages = {2733–2743},
numpages = {11},
location = {Taipei, Taiwan},
series = {SIGIR '23}
}

@inproceedings{pelrine2023towards,
    title = "Towards Reliable Misinformation Mitigation: Generalization, Uncertainty, and {GPT}-4",
    author = "Pelrine, Kellin  and
      Imouza, Anne  and
      Thibault, Camille  and
      Reksoprodjo, Meilina  and
      Gupta, Caleb  and
      Christoph, Joel  and
      Godbout, Jean-Fran{\c{c}}ois  and
      Rabbany, Reihaneh",
    editor = "Bouamor, Houda  and
      Pino, Juan  and
      Bali, Kalika",
    booktitle = "Proceedings of the 2023 Conference on Empirical Methods in Natural Language Processing",
    month = dec,
    year = "2023",
    address = "Singapore",
    publisher = "Association for Computational Linguistics",
    url = "https://aclanthology.org/2023.emnlp-main.395/",
    doi = "10.18653/v1/2023.emnlp-main.395",
    pages = "6399--6429"
}

@inbook{bohan2024disinformation,
author = {Bohan Jiang and Zhen Tan and Ayushi Nirmal and Huan Liu},
title = {Disinformation Detection: An Evolving Challenge in the Age of LLMs},
booktitle = {Proceedings of the 2024 SIAM International Conference on Data Mining (SDM)},
chapter = {},
pages = {427-435},
doi = {10.1137/1.9781611978032.50},
URL = {https://epubs.siam.org/doi/abs/10.1137/1.9781611978032.50},
eprint = {https://epubs.siam.org/doi/pdf/10.1137/1.9781611978032.50},
    year = {2024}
}

@inproceedings{tariq2025p2e, series={MM ’25},
   title={From Prediction to Explanation: Multimodal, Explainable, and Interactive Deepfake Detection Framework for Non-Expert Users},
   url={http://dx.doi.org/10.1145/3746027.3755786},
   DOI={10.1145/3746027.3755786},
   booktitle={Proceedings of the 33rd ACM International Conference on Multimedia},
   publisher={ACM},
   author={Tariq, Shahroz and Woo, Simon S. and Singh, Priyanka and Irmalasari, Irena and Gupta, Saakshi and Gupta, Dev},
   year={2025},
   month=oct, pages={11716–11725},
   collection={MM ’25} }

@misc{nakamura2020rfakeddit,
      title={r/Fakeddit: A New Multimodal Benchmark Dataset for Fine-grained Fake News Detection}, 
      author={Kai Nakamura and Sharon Levy and William Yang Wang},
      year={2020},
      eprint={1911.03854},
      archivePrefix={arXiv},
      primaryClass={cs.CL},
      url={https://arxiv.org/abs/1911.03854}, 
}

@misc{papadopoulos2023verite,
      title={VERITE: A Robust Benchmark for Multimodal Misinformation Detection Accounting for Unimodal Bias}, 
      author={Stefanos-Iordanis Papadopoulos and Christos Koutlis and Symeon Papadopoulos and Panagiotis C. Petrantonakis},
      year={2023},
      eprint={2304.14133},
      archivePrefix={arXiv},
      primaryClass={cs.CV},
      url={https://arxiv.org/abs/2304.14133}, 
}

@misc{singh2025dgm4ext,
      title={DGM4+: Dataset Extension for Global Scene Inconsistency}, 
      author={Gagandeep Singh and Samudi Amarsinghe and Priyanka Singh and Xue Li},
      year={2025},
      eprint={2509.26047},
      archivePrefix={arXiv},
      primaryClass={cs.CV},
      url={https://arxiv.org/abs/2509.26047}, 
}

@misc{liu2023llava,
      title={Visual Instruction Tuning}, 
      author={Haotian Liu and Chunyuan Li and Qingyang Wu and Yong Jae Lee},
      year={2023},
      eprint={2304.08485},
      archivePrefix={arXiv},
      primaryClass={cs.CV},
      url={https://arxiv.org/abs/2304.08485}, 
}

@misc{meta2024llama3,
  title={meta-llama/Llama-3.2-11B-Vision-Instruct},
  author={Meta},
  howpublished={\url{https://huggingface.co/meta-llama/Llama-3.2-11B-Vision-Instruct/}},
  year={2024}
}

@misc{serper,
  title={Serper API},
  author={Serper},
  howpublished={\url{https://serper.dev/}},
}

@misc{googlevision,
  title={Google Vision API},
  author={Google},
  howpublished={\url{https://cloud.google.com/vision?hl=en}},
}

@article{park2025genai,
  title={Generative AI and misinformation: a scoping review of the role of generative AI in the generation, detection, mitigation, and impact of misinformation},
  author={Park, Seyeon and Nan, Xiaoli},
  journal={AI \& SOCIETY},
  pages={1--15},
  year={2025},
  publisher={Springer}
}

@article{massenon2025myphone,
  title={” My AI is Lying to Me”: User-reported LLM hallucinations in AI mobile apps reviews},
  author={Massenon, Rhodes and Gambo, Ishaya and Khan, Javed Ali and Agbonkhese, Christopher and Alwadain, Ayed},
  journal={Scientific Reports},
  volume={15},
  number={1},
  pages={30397},
  year={2025},
  publisher={Nature Publishing Group UK London}
}

@article{touvron2023llama,
  title={Llama: Open and efficient foundation language models},
  author={Touvron, Hugo and Lavril, Thibaut and Izacard, Gautier and Martinet, Xavier and Lachaux, Marie-Anne and Lacroix, Timoth{\'e}e and Rozi{\`e}re, Baptiste and Goyal, Naman and Hambro, Eric and Azhar, Faisal and others},
  journal={arXiv preprint arXiv:2302.13971},
  year={2023}
}

@article{team2023gemini,
  title={Gemini: a family of highly capable multimodal models},
  author={Team, Gemini and Anil, Rohan and Borgeaud, Sebastian and Alayrac, Jean-Baptiste and Yu, Jiahui and Soricut, Radu and Schalkwyk, Johan and Dai, Andrew M and Hauth, Anja and Millican, Katie and others},
  journal={arXiv preprint arXiv:2312.11805},
  year={2023}
}

@article{huang2025survey,
author = {Huang, Lei and Yu, Weijiang and Ma, Weitao and Zhong, Weihong and Feng, Zhangyin and Wang, Haotian and Chen, Qianglong and Peng, Weihua and Feng, Xiaocheng and Qin, Bing and Liu, Ting},
title = {A Survey on Hallucination in Large Language Models: Principles, Taxonomy, Challenges, and Open Questions},
year = {2025},
issue_date = {March 2025},
publisher = {Association for Computing Machinery},
address = {New York, NY, USA},
volume = {43},
number = {2},
issn = {1046-8188},
url = {https://doi.org/10.1145/3703155},
doi = {10.1145/3703155},
journal = {ACM Trans. Inf. Syst.},
month = jan,
articleno = {42},
numpages = {55}
}

@misc{huang2023reasoning,
      title={Towards Reasoning in Large Language Models: A Survey}, 
      author={Jie Huang and Kevin Chen-Chuan Chang},
      year={2023},
      eprint={2212.10403},
      archivePrefix={arXiv},
      primaryClass={cs.CL},
      url={https://arxiv.org/abs/2212.10403}, 
}

@article{costello2024dialogue,
author = {Thomas H. Costello  and Gordon Pennycook  and David G. Rand },
title = {Durably reducing conspiracy beliefs through dialogues with AI},
journal = {Science},
volume = {385},
number = {6714},
pages = {eadq1814},
year = {2024},
doi = {10.1126/science.adq1814},
URL = {https://www.science.org/doi/abs/10.1126/science.adq1814},
eprint = {https://www.science.org/doi/pdf/10.1126/science.adq1814},
}

@inproceedings{cepeda2023geoclip,
title={Geo{CLIP}: Clip-Inspired Alignment between Locations and Images for Effective Worldwide Geo-localization},
author={Vicente Vivanco Cepeda and Gaurav Kumar Nayak and Mubarak Shah},
booktitle={Thirty-seventh Conference on Neural Information Processing Systems},
year={2023},
url={https://openreview.net/forum?id=I18BXotQ7j}
}

@article{nan2022publichealth,
author = {Xiaoli Nan and Irina A. Iles and Bo Yang and Zexin Ma},
title = {Public Health Messaging during the COVID-19 Pandemic and Beyond: Lessons from Communication Science},
journal = {Health Communication},
volume = {37},
number = {1},
pages = {1--19},
year = {2022},
publisher = {Routledge},
doi = {10.1080/10410236.2021.1994910},
note ={PMID: 34724838},
URL = {https://doi.org/10.1080/10410236.2021.1994910},
eprint = {https://doi.org/10.1080/10410236.2021.1994910}
}

@misc{openai2023gpt4,
title={GPT‑4},
author={OpenAI}
}

@article{meel2020fakenews,
title = {Fake news, rumor, information pollution in social media and web: A contemporary survey of state-of-the-arts, challenges and opportunities},
journal = {Expert Systems with Applications},
volume = {153},
pages = {112986},
year = {2020},
issn = {0957-4174},
doi = {https://doi.org/10.1016/j.eswa.2019.112986},
url = {https://www.sciencedirect.com/science/article/pii/S0957417419307043},
author = {Priyanka Meel and Dinesh Kumar Vishwakarma},
}

@article{sun2024ai,
  title={AI hallucination: towards a comprehensive classification of distorted information in artificial intelligence-generated content},
  author={Sun, Yujie and Sheng, Dongfang and Zhou, Zihan and Wu, Yifei},
  journal={Humanities and Social Sciences Communications},
  volume={11},
  number={1},
  pages={1--14},
  year={2024},
  publisher={Palgrave}
}

@article{hameleers2020picture,
author = {Michael Hameleers and Thomas E. Powell and Toni G.L.A. Van Der Meer and Lieke Bos},
title = {A Picture Paints a Thousand Lies? The Effects and Mechanisms of Multimodal Disinformation and Rebuttals Disseminated via Social Media},
journal = {Political Communication},
volume = {37},
number = {2},
pages = {281--301},
year = {2020},
publisher = {Routledge},
doi = {10.1080/10584609.2019.1674979},
URL = {https://doi.org/10.1080/10584609.2019.1674979},
eprint = {https://doi.org/10.1080/10584609.2019.1674979}
}

\end{document}